\documentclass[10pt,twocolumn,letterpaper]{article}

\usepackage{iccv}
\usepackage{times}
\usepackage{epsfig}
\usepackage{graphicx}
\usepackage{amsmath}
\usepackage{amssymb}

\usepackage{comment}
\usepackage{booktabs}
\usepackage{multirow}
\usepackage{multicol}
\usepackage{xcolor}
\usepackage{adjustbox}
\usepackage{import}
\usepackage{color}
\usepackage{url}            
\usepackage{amsfonts}       
\usepackage{nicefrac}       
\usepackage{microtype}      
\usepackage{csquotes}
\usepackage{subfig}
\usepackage{array}
\usepackage{adjustbox}
\usepackage{xspace}
\usepackage{wrapfig}

\definecolor{mediumpersianblue}{rgb}{0.2, 0.4, 0.8}
\newcommand{\perminus}[1]{{\color{blue}{\small\bf (-#1)}}}

\iccvfinalcopy 

\ificcvfinal\fi

\graphicspath{{ICCV2021/Arxiv/}}

\usepackage[pagebackref=true,breaklinks=true,colorlinks,bookmarks=false]{hyperref}

\def\ours{\texttt{AdaMML}\xspace}
\DeclareMathOperator*{\argmax}{arg\,max}
\DeclareMathOperator{\E}{\mathbb{E}}

\begin{document}

\title{AdaMML: Adaptive Multi-Modal Learning for Efficient Video Recognition}

\author{Rameswar Panda$^{1,\dagger}$, Chun-Fu (Richard) Chen$^{1,\dagger}$, Quanfu Fan$^{1}$, Ximeng Sun$^{2}$, \\ Kate Saenko$^{1,2}$, Aude Oliva$^{1,3}$, Rogerio Feris$^{1}$ \\
$\dagger$: Equal Contribution \\
$^1$MIT-IBM Watson AI Lab, $^2$Boston University, $^3$MIT  
}

\maketitle
\ificcvfinal\thispagestyle{empty}\fi

\begin{abstract}

Multi-modal learning, which focuses on utilizing various modalities to improve the performance of a model, is widely used in video recognition. While traditional multi-modal learning offers excellent recognition results, its computational expense limits its impact for many real-world applications. In this paper, we propose an adaptive multi-modal learning framework, called AdaMML, that selects on-the-fly the optimal modalities for each segment conditioned on the input for efficient video recognition. Specifically, given a video segment, a multi-modal policy network is used to decide what modalities should be used for processing by the recognition model, with the goal of improving both accuracy and efficiency. We efficiently train the policy network jointly with the recognition model using standard back-propagation. Extensive experiments on four challenging diverse datasets demonstrate that our proposed adaptive approach yields $35\%-55\%$ reduction in computation when compared to the traditional baseline that simply uses all the modalities irrespective of the input, while also achieving consistent improvements in accuracy over the state-of-the-art methods.

\end{abstract}

\section{Introduction}
\label{sec:introduction}

\begin{figure*}[t]
\begin{center}
     \includegraphics[width=\linewidth]{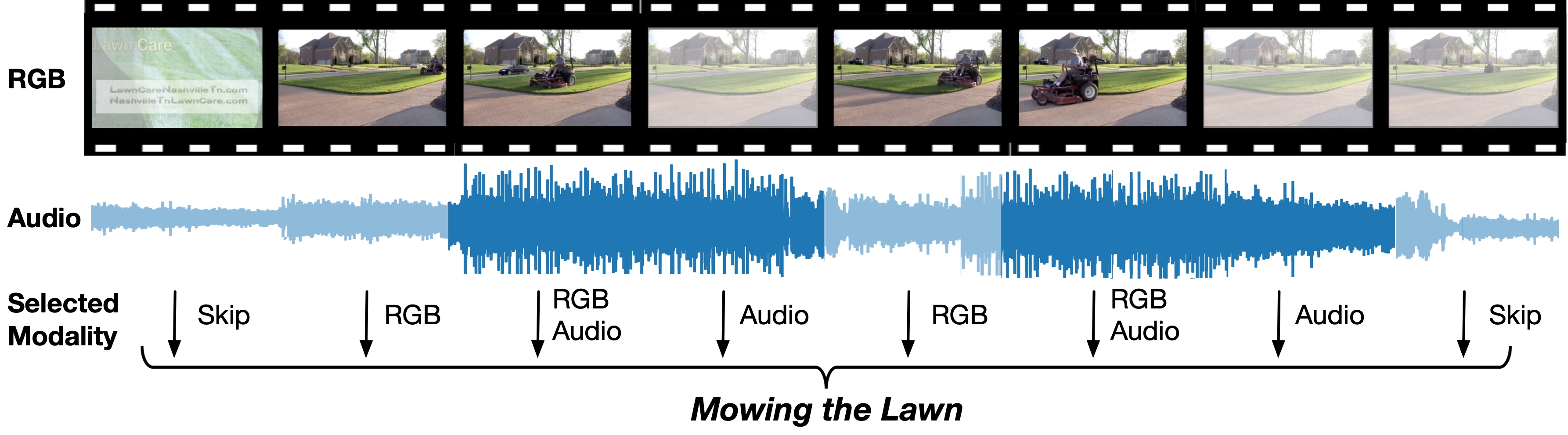}
\end{center} 
  \vspace{-6mm}
   \caption{
   \small
   \textbf{A conceptual overview of our approach}. Rather than processing both RGB and Audio modalities for all the segments, our approach learns a policy to select the optimal modalities per input segment, that is needed to correctly recognize an action in a given video. In the figure, the lawn mower is moving with relevant audio only in the third and sixth segment, therefore those segments could be processed using both modalities, while the rest of the segments require only one modality (e.g., only audio is relevant for the fourth segment as the lawn mower moves outside of the camera but its sound is still clear) or even skipped (e.g., both of the modalities are irrelevant in the first and the last segment), without losing any accuracy. Note that our approach can be extended to any number of modalities as shown in experiments. 
   }
   \vspace{-5pt}
   \label{fig:teaser}
\end{figure*}

Videos are rich in multiple modalities: RGB frames, motion (optical flow), and audio. As a result, multi-modal learning which focuses on utilizing various modalities to improve the performance of a video recognition model, has attracted much attention in the recent years. 
Despite encouraging progress, multi-modal learning becomes computationally impractical in real-world scenarios where the videos are untrimmed and span several minutes or even hours. Given a long video, some modalities often provide irrelevant/redundant information for the recognition of the action class. Thus, utilizing information from all the input modalities may be counterproductive as informative modalities are often overwhelmed by uninformative ones in long videos.
Furthermore, some modalities require more computation than others and hence selecting the cheaper modality with good performance can significantly save computation leading to more efficient video recognition. 

Let us consider the video in Figure~\ref{fig:teaser}, represented by eight uniformly sampled video segments from a video. We ask, \textit{Do all the segments require both RGB and audio stream to recognize the action as \enquote{Mowing the Lawn} in this video?} The answer is clear: No, the lawn mower is moving with relevant audio only in the third and sixth segment, therefore we need both RGB and audio streams for these two video segments to improve the model confidence for recognizing the correct action, while the rest of the segments can be processed with only one modality or even skipped (e.g., the first and last video segment) without losing any accuracy, resulting in large computational savings compared to processing all the segments using both modalities.
Thus, in contrast to the commonly used one-size-fits-all scheme for multi-modal learning, we would like these decisions to be made individually per input segment, leading to different amounts of computation for different videos. Based on this intuition, we present a new perspective for efficient video recognition by adaptively selecting input modalities, on a per segment basis, for recognizing complex actions.

In this paper, we propose \ours, a novel and differentiable approach to learn a decision policy that selects optimal modalities conditioned on the inputs for efficient video recognition. Specifically, our main idea is to learn a model (referred to as the multi-modal policy network) that outputs the posterior probabilities of all the binary decisions for using or skipping each modality on a per segment basis.
As these decision functions are discrete and non-differentiable, we rely on an efficient Gumbel-Softmax sampling approach~\cite{jang2016categorical} to learn the decision policy jointly with the network parameters through standard back-propagation, without resorting to complex reinforcement learning as in~\cite{wu2019adaframe,yeung2016end}.
We design the objective function to achieve both competitive performance and efficiency required for video recognition. We demonstrate that adaptively selecting input modalities by a lightweight policy network yields not only significant savings in computation (e.g., about $47.3\%$ and $35.2\%$ less GFLOPS compared to a weighted fusion baseline that simply uses all the modalities, on Kinetics-Sounds~\cite{arandjelovic2017look} and ActivityNet~\cite{caba2015activitynet}, respectively), but also consistent improvement in accuracy the over state-of-the-art methods.

The main contributions of our work are as follows:
\vspace{-2mm}
\begin{itemize} \setlength{\itemsep}{-0.2pt}
    \item We propose a novel and differentiable approach that automatically determines what modalities to use per segment per input for efficient video recognition. This is in sharp contrast to current multi-modal learning approaches that utilizes all the input modalities without considering their relevance to the video recognition. 
    \item We efficiently train the multi-modal policy network jointly with the recognition model using standard back-propagation through Gumbel-Softmax sampling. 
    \item We conduct extensive experiments on four video benchmarks (Kinetics-Sounds~\cite{arandjelovic2017look}, ActivityNet~\cite{caba2015activitynet}, FCVID~\cite{jiang2017exploiting} and Mini-Sports1M~\cite{karpathy2014large}) with different multi-modal learning tasks (RGB + Audio, RGB + Flow, and RGB + Flow + Audio) to demonstrate the superiority of our approach over state-of-the-art methods.
\end{itemize}

\section{Related Work}
\label{sec:relatedwork}

Our work relates to three major research directions: efficient video recognition, multi-modal learning and adaptive computation. Here, we focus on some representative methods closely related to our work.

\vspace{1mm}
\noindent\textbf{Efficient Video Recognition.} Video recognition has been one of the most active research areas in computer vision recently. In the context of deep neural networks, it is typically performed by either 2D-CNNs~\cite{karpathy2014large,wang2016temporal,fan2019more,wang2018non,fan2019more,lin2019tsm,zhou2018temporal} or 3D-CNNs~\cite{tran2015learning,carreira2017quo,hara2018can,feichtenhofer2020x3d}.
While extensive studies have been conducted in the last few years, limited efforts have been made towards efficient video recognition. Specifically, methods for efficient recognition focus on either designing new lightweight architectures (e.g., Tiny Video Networks~\cite{piergiovanni2019tiny}, channel-separated CNNs~\cite{tran2019video}, and X3D~\cite{feichtenhofer2020x3d}) or selecting salient frames/clips~\cite{yeung2016end,wu2019adaframe,korbar2019scsampler,gao2020listentolook,wu2019multi,hussein2020timegate,meng2020ar}. 
Our approach is most related to the latter which focuses on conditional computation for videos and is agnostic to the network architecture used for recognizing videos. Representative methods typically use reinforcement learning (RL) policy gradients~\cite{yeung2016end,wu2019adaframe} or audio~\cite{korbar2019scsampler,gao2020listentolook} to select relevant video frames. Recently, LiteEval~\cite{wu2019liteeval} proposes a coarse-to-fine framework that uses a binary gate for selecting either coarse or fine features.
Unlike existing works, our proposed approach focuses on the multi-modal nature of videos and adaptively selects the right modality per input instance for recognizing complex actions in long videos.   
Moreover, our framework is fully differentiable, and thus is easier to train than complex RL policy gradients~\cite{yeung2016end,wu2019adaframe,wu2019multi}.

\begin{figure*}[bt]
\centering
    \includegraphics[width=\linewidth]{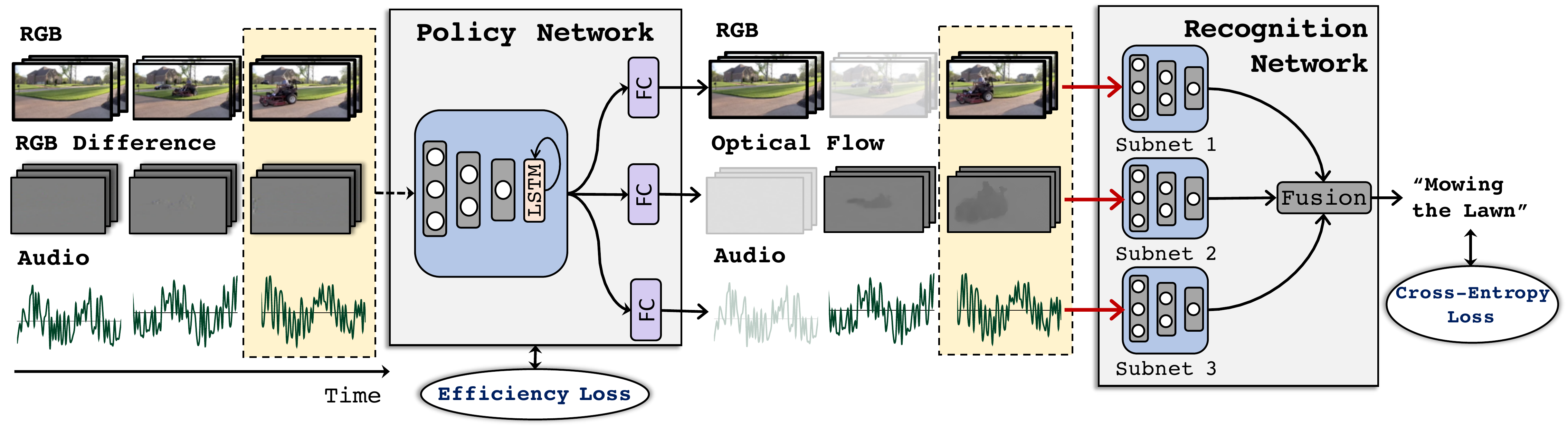}
  \vspace{-5mm}
   \caption{ \small
   \textbf{Illustration of our approach}. \ours consists of a lightweight policy network and a recognition network composed of different sub-networks that are trained jointly (via late fusion with learnable weights) for recognizing videos. The policy network decides what modalities to use on a per segment basis to achieve the best recognition accuracy and efficiency in video recognition. 
   In training, policies are sampled from a Gumbel-Softmax distribution, which allows us to optimize the policy network via backpropagation. During inference, an input segment is first fed into the policy network and then selected modalities are routed to the recognition network to generate segment-level predictions. Finally, the network averages all the segment-level predictions to obtain the video-level prediction.
   Best viewed in color.
   }
   \vspace{-5pt}
   \label{fig:model_overview}
\end{figure*}

\noindent\textbf{Multi-Modal Learning.} Multi-modal learning has been studied from multiple perspectives, such as two stream networks that fuse decisions from multiple modalities for classification~\cite{simonyan2014two,carreira2017quo,kazakos2019epic,kiela2018efficient,arevalo2017gated}, and cross-modal learning that takes one modality as input and make prediction on the other modality~\cite{korbar2018cooperative,arandjelovic2017look,zhao2018sound,antol2015vqa,frome2013devise,socher2013zero}. Recent work in~\cite{wang2019makes} addresses the problem of joint training in multi-modal networks, without deciding which modality to focus for a given input sample as in our current approach. Our proposed \ours framework is also related to prior works in joint appearance and motion modeling~\cite{sun2018optical,lee2018motion,crasto2019mars} that focuses on combining RGB and optical flow streams. Design of different fusion schemes~\cite{perez2019mfas} through neural architecture search~\cite{zoph2016neural} is also another recent trend for multi-modal learning. In contrast, we propose an instance-specific general framework for automatically selecting the right modality per segment for efficient video recognition. 

\noindent\textbf{Adaptive Computation.} Many adaptive computation methods have been recently proposed with the goal of improving computational efficiency~\cite{bengio2015conditional,bengio2013estimating,veit2018convolutional,wang2018skipnet,graves2016adaptive,figurnov2017spatially,mcgill2017deciding,meng2020ar}. While BlockDrop~\cite{wu2018blockdrop} dynamically selects which layers to execute per sample during inference, GaterNet~\cite{chen2019you} proposes a gating network to learn channel-wise binary gates for the main network. Channel gating network~\cite{hua2019channel} identifies regions in the features that contribute less to the classification result, and skips the computation on a subset of the input channels for these ineffective regions. SpotTune~\cite{guo2019spottune} learns to adaptively route information through fine-tuned or pre-trained layers for different tasks. 
Adaptive selection of different regions for fast object detection is presented in~\cite{najibi2019autofocus,gao2018dynamic}.
While our approach is inspired by these methods, in this paper, our goal is to adaptively select optimal modalities per input instance to improve efficiency in video recognition.  
To the best of our knowledge, this is the first work on data-dependent selection of different modalities for efficient video recognition.

\section{Proposed Method}
\label{sec:proposedmethod}

Given a video $V$ containing a sequence of segments $\{s_1, s_2, \cdots, s_T \}$ over $K$ input modalities $\{\mathcal{M}_1, \mathcal{M}_2, \cdots, \mathcal{M}_K\}$, our goal is to seek an adaptive multi-modal selection policy that decides what input modalities should be used for each segment in order to improve the accuracy, while taking the computational efficiency into account for video recognition. 

\subsection{Approach Overview} \label{sec:overview}
Figure~\ref{fig:model_overview} illustrates an overview of our approach. Treating the task of finding an optimal multi-modal selection policy as a search problem quickly becomes intractable as the number of potential configurations grows exponentially with the number of video segments and modalities. 
Instead of handcrafting the selections, we develop a policy network that contains a very lightweight joint feature extractor and an LSTM module to output a binary policy vector per segment per input, representing whether to keep or drop an input modality for efficient multi-modal learning.

During training, the policy network is jointly trained with the recognition network using Gumbel-Softmax sampling~\cite{jang2016categorical}. At test time, first an input video segment is fed into the policy network, whose output decides the right modalities to use for the given segment, and then the selected input modalities are routed to the corresponding sub-networks in the recognition network to generate the segment-level predictions. Finally, the network averages all the segment-level predictions as the video-level prediction. Note that the additional computational cost incurred by the lightweight policy network (MobileNetv2~\cite{sandler2018mobilenetv2} in our case) is negligible in comparison to the recognition model. 

\subsection{Learning Adaptive Multi-Modal Policy}
\label{sec:policy}

\noindent\textbf{Multi-Modal Policy Network.} The policy network contains a lightweight joint feature extractor and an LSTM module for modeling the causality across different time steps in a video. Specifically, at the $t$-th time step, the LSTM takes in the joint feature $f_t$ of the current video segment $s_t$, previous hidden states $h_{t-1}$, cell outputs $o_{t-1}$ to compute the current hidden state $h_t$ and cell states $o_t$:
\begin{equation}
    \label{eq:lstm}
    h_t, o_t = \text{LSTM}(f_t, h_{t-1}, o_{t-1}).
\end{equation}
Given the hidden state, the policy network estimates a policy distribution for each modality and samples binary decisions $u_{t,k}$ indicating whether to select modality $k$ at time step $t$ ($\mathbf{U} = \{\mathbf{u}_{t,k}\}_{l\le T, k\le K}$) via Gumbel-Softmax operation as described next. Given the decisions, we forward the current segment to corresponding sub-networks to get a segment-level prediction and average all segment-level predictions to generate video-level prediction for an input video.

\vspace{1mm}
\noindent\textbf{Training using Gumbel-Softmax Sampling.} \ours makes decisions about skipping or using each modality per segment per input. However, the fact that the policy is discrete makes the network non-differentiable and therefore difficult to be optimized with standard backpropagation.
One way to solve this is to convert the optimization to a reinforcement learning problem and then derive the optimal parameters of the policy network with policy gradient methods~\cite{williams1992simple,sutton2018reinforcement}. However, RL policy gradients are often complex, unwieldy to train and require techniques to reduce variance during training as well as it is slow to converge in many applications~\cite{wu2018blockdrop,wu2019liteeval,jang2016categorical,wu2019multi}.
As an alternative, in this paper, we adopt Gumbel-Softmax sampling~\cite{jang2016categorical} to resolve this non-differentiability and enable direct optimization of the discrete policy in an efficient way.

The Gumbel-Softmax trick~\cite{jang2016categorical} is a simple and effective way to replace the original non-differentiable sample from a discrete distribution with a differentiable sample from a corresponding Gumbel-Softmax distribution.  
Specifically, at each time step $t$, we first generate the logits $z_k\in\mathbb{R}^2$ (i.e, output scores of policy network for modality $k$) from hidden states $h_t$ by a fully-connected layer $z_k={FC}_k(h_t,{\theta_{FC}}_k)$ for each modality and then use the Gumbel-Max trick \cite{jang2016categorical} to draw discrete samples from a categorical distribution as: 
\begin{equation} 
\label{eq:gumbelmax}
    \hat{P}_k = \argmax_{i\in \{0,1\}} (\log z_{i,k}+G_{i,k}), \ \ \ \ \ k \in [1, ..., K] \vspace{-1mm}
\end{equation} 
where $G_{i,k}=-\log(-\log U_{i,k})$ is a standard Gumbel distribution with $U_{i,k}$ sampled from a uniform i.i.d  distribution $Unif(0,1)$.
Due to non-differentiable property of $\argmax$ operation in Equation~\ref{eq:gumbelmax}, Gumbel-Softmax distribution \cite{jang2016categorical} is thus used as a continuous relaxation to $\operatorname*{arg\,max}$. Accordingly, sampling from a Gumbel-Softmax distribution allows us to backpropagate from discrete samples to the policy network. We represent $\hat{P}_{k}$ as a one-hot vector and then one-hot coding is relaxed to a real-valued vector ${P}_k$ using softmax: 
\begin{equation}
\label{eq:one}
    P_{i,k}=\frac{\exp((\log z_{i,k}+G_{i,k})/\tau)}{\sum_{j\in \{0,1\}} \exp((\log z_{j,k}+G_{j,k})/\tau)}, 
\end{equation}
where $i\in \{0,1\}, \ k \in [1, ..., K]$, $\tau$ is a temperature parameter, which controls the discreteness of $P_{k}$, as $\lim\limits_{\tau\to +\infty} P_k$ converges to a uniform distribution and $\lim\limits_{\tau\to 0} P_k$ becomes a one-hot vector.
More specifically, when $\tau$ becomes closer to $0$, the samples from the Gumbel Softmax distribution become indistinguishable from the discrete distribution (i.e, almost the same as the one-hot vector). 
In summary, during the forward pass, we sample the policy using Equation~\ref{eq:gumbelmax} and during the backward pass, we approximate the gradient of the discrete samples by computing the gradient of the continuous softmax relaxation in Equation~\ref{eq:one}. 

\subsection{Loss Function} \label{sec:loss}
Let $\Theta=\{\theta_{\Phi},\theta_{LSTM},\theta_{{FC}_1},...,\theta_{{FC}_{K}},$
$\theta_{\Psi_1},...,\theta_{\Psi_{K}}\}$ denote the total trainable parameters in our framework, where $\theta_{\Phi}$ and $\theta_{LSTM}$ represent the parameters of the joint feature extractor and LSTM used in the policy network respectively. $\theta_{{FC}_1},...,\theta_{{FC}_{K}}$ represent the parameters of the fully connected layers that generate policy logits from the LSTM hidden states and $\theta_{\Psi_1},...,\theta_{\Psi_{K}}$ represent the parameters of $K$ sub-networks that are jointly trained for recognizing video. During training, we minimize the following loss to encourage both correct predictions as well as minimize the selection of modalities that require more computation. 
\begin{equation}
\begin{split}
\label{eq:loss}
    \E_{(V,y)\sim \mathcal{D}_{train}}\left[-y\log(\mathcal{P}(V; \Theta)) + \sum\limits_{k=1}^{K} \lambda_k \mathcal{C}_k\right], \\ \ \ \ \mathcal{C}_k = \left\{ \begin{array}{ll}
    (\dfrac{|U_k|_0}{C})^2 & \text{if correct} \\
    \gamma & \text{otherwise}
  \end{array}
  \right.
\end{split}
\end{equation}
where $\mathcal{P}(V; \Theta)$ and $y$ represents the prediction and one-hot encoded ground truth label of the training video sample $V$ and $\lambda_k$ represents the cost associated with processing $k$-th modality. $U_k$ represents the decision policy for $k$-th modality and $\mathcal{C}_k=(\dfrac{|U_k|_0}{C})^2$ measures the fraction of segments that selected modality $k$ out of total $C$ video segments; when a correct prediction is produced. We penalize incorrect predictions with $\gamma$, which including $\lambda_k$ controls the trade-off between efficiency and accuracy. We use these parameters to vary the operating point of our model, allowing different models to be trained depending on the target budget constraint. While the first part of the Equation~\ref{eq:loss} represents the standard cross-entropy loss to measure the classification quality, the second part drives the network to learn a policy that favors selection of modality that is computationally more efficient in recognizing videos (e.g., processing RGB frames requires more computation than the audio streams). 

\section{Experiments}
\label{sec:experiments}
In this section, we conduct extensive experiments on four standard datasets to show that \ours outperforms many strong baselines including state-of-the-art methods while significantly reducing computation and qualitative analysis to verify the effectiveness of our adaptive policy learning.

\begin{table*}[t]
\centering
\begin{adjustbox}{max width=\textwidth}
\begin{tabular}{ c|c|c|c|c || c|c|c|c}
\toprule
Dataset & \multicolumn{4}{c ||}{Kinetics-Sounds} &  \multicolumn{4}{c}{ActivityNet} \\
 \cmidrule{2-9}
 &  & \multicolumn{2}{c|}{Selection Rate (\%)} &   &  &  \multicolumn{2}{c|}{Selection Rate (\%)} &  \\
 Method      &   Acc. (\%)  &   RGB & Audio    & GFLOPs &  mAP (\%)  &   RGB & Audio    & GFLOPs \\
\midrule
RGB      & 82.85 &  100  &  $-$  & 141.36   & 73.24 &  100  &  $-$  & 141.36 \\
Audio    & 65.49 &  $-$  &  100  & 3.82   & 13.88 &  $-$  &  100  & 3.82   \\
\midrule
Weighted Fusion  & 87.86  &  100  &  100  & 145.17  & 72.88 &  100  &  100  & 145.17 \\
\ours      & \textbf{88.17} & 46.47  & 94.15  & \textbf{76.45} \perminus{47.3\%}  & \textbf{73.91} & 76.25 & 56.35 & \textbf{94.01} \perminus{35.2\%} \\
\bottomrule
\end{tabular}
\end{adjustbox} \vspace{-1mm}
\caption{\small \textbf{Video recognition results with RGB + Audio modalities on Kinetics-Sounds and ActivityNet}. On both datasets, our proposed approach \ours outperforms the weighted fusion baseline while offering significant computational savings (shown in \textcolor{blue}{blue}). 
} 
\label{tab:afm} 
\end{table*}
\begin{table}[!th]
\centering
\begin{adjustbox}{max width=\linewidth}
\begin{tabular}{ c|c|c|c|c}
\toprule
&  & \multicolumn{2}{c|}{Selection Rate (\%)} &   \\
Method &  Acc. (\%)  &   RGB & Flow    &  GFLOPs  \\
\midrule
RGB         & 82.85 &  100  &  $-$  & 141.36 \\
Flow      & 75.73 &  $-$  &  100  &  163.39  \\
\midrule
Weighted Fusion   & 83.47  &  100  &  100  & 304.75 \\
\ours-\texttt{Flow}        & 83.82 & 56.04 & 36.39 & 151.54 \perminus{50.3\%} \\
\ours-\texttt{RGBDiff}       & \textbf{84.36} & 44.61 & 37.40 & \textbf{137.03} \perminus{55.0\%} \\
\bottomrule
\end{tabular} 
\end{adjustbox} \vspace{-1mm}
\caption{\small \textbf{RGB + Flow on Kinetics-Sounds}. \ours-\texttt{RGBDiff} obtains best performance with more than $50\%$ savings in GFLOPs.}
\label{tab:rgb_flow_k}
\vspace{-1pt}
\end{table}

\begin{table}[!th]
\centering
\begin{adjustbox}{max width=\linewidth}
\begin{tabular}{ c|c|c|c|c|c}
\toprule
 &  & \multicolumn{3}{c|}{Selection Rate (\%)} &   \\
Method &  Acc. (\%)   &   RGB & Flow & Audio   & GFLOPs   \\
\midrule
RGB         & 82.85  &  100   & $-$ &  $-$ & 141.36 \\
Flow      & 75.73 &  $-$  & 100 & $-$ & 163.39   \\
Audio       & 65.49 &  $-$  & $-$ &  100 & 3.82    \\
\midrule
Weighted Fusion   & 88.25 &  100  &  100 & 100 & 308.56  \\
\ours-\texttt{Flow}       & 88.54 & 56.13 & 20.31 & 97.49 & \textbf{132.94} \perminus{56.9\%} \\
\ours-\texttt{RGBDiff}       & \textbf{89.06} & 55.06 & 26.82 & 95.12  & 141.97 \perminus{54.0\%} \\
\bottomrule
\end{tabular}
\end{adjustbox} \vspace{-1mm}
\caption{\small \textbf{RGB + Flow + Audio on Kinetics-Sounds}. \ours-\texttt{RGBDiff} obtains the best accuracy of $89.06\%$ which is $6.21\%$ more than RGB only performance with similar GFLOPS.} 
\label{tab:rgb_audio_flow_k}
\vspace{-2pt}
\end{table}

\subsection{Experimental Setup}

\noindent\textbf{Datasets and Tasks.}
We evaluate the performance of our approach using four datasets, namely Kinetics-Sounds~\cite{arandjelovic2017look}, ActivityNet-v1.3~\cite{caba2015activitynet}, FCVID~\cite{jiang2017exploiting}, and Mini-Sports1M~\cite{karpathy2014large}.
Kinetics-Sounds is a subset of Kinetics~\cite{carreira2017quo} and consists of $22,521$ videos for training and $1,532$ videos testing across $31$ action classes~\cite{gao2020listentolook}\footnote{The Kinetics-Sounds dataset assembled by~\cite{arandjelovic2017look} consists of 34 classes. However, 3 classes were removed from the original Kinetics dataset. Hence, we use the remaining 31 classes in our experiments, as in~\cite{gao2020listentolook}.}.
ActivityNet contains $10,024$ videos for training and $4,926$ videos for validation across $200$ action categories. FCVID has $45,611$ videos for training and $45,612$ videos for testing across $239$ classes. Mini-Sports1M (assembled by~\cite{gao2020listentolook}) is a subset of full Sports1M dataset~\cite{karpathy2014large} containing $30$ videos per class in training and $10$ videos per class in testing with a total of $487$ action classes.
We consider three groups of multi-modal learning tasks such as (I) RGB + Audio, (II) RGB + Flow, and (III) RGB + Flow + Audio on different datasets. More details about the datasets can be found in the appendix. 

\vspace{1mm}
\noindent\textbf {Data Inputs.}
For each input segment, we take around 1-second of data and temporally align all the modalities. For RGB, we uniformly sample $8$ frames out of $32$ consecutive frames ($8\times224\times224$); and for optical flow, we stack $10$ interleaved horizontal and vertical optical flow frames~\cite{wang2016temporal}. For audio, we use a 1-channel audio-spectrogram as input~\cite{kazakos2019epic} ($256\times256$, which is 1.28 seconds audio segment).
Note that since computing optical flow is very expensive, we utilize RGB frame difference as a proxy to flow in our policy network and compute flow when needed.
For RGB frame difference, we follow similar approach used in optical flow and use an input clip $15\times8\times224\times224$ by simply computing the frame differences.  
For the policy network, we further subsample the input data for non-audio modality, e.g., the RGB input becomes $4\times160\times160$.

\vspace{1mm}
\noindent\textbf{Implementation Details.} 
For the recognition network, we use TSN-like ResNet-50~\cite{wang2016temporal} for both RGB and Flow modalities, and MobileNetV2~\cite{sandler2018mobilenetv2} for the audio modality. We simply apply late-fusion with learnable weights over the predictions from each modality to obtain the final prediction. We use MobileNetV2 for all modalities in the policy network to extract features and then apply two additional FC layers with dimension $2,048$ to concatenate the features from all modalities as the joint-feature. 
The hidden dimension of LSTM is set to $256$. We use $K$ parallel FC layers on top of LSTM outputs to generate the binary decision policy for each modality. The computational cost for processing RGB + Audio in the policy network and the recognition network are $0.76$ and $14.52$ GFLOPs, respectively.

\vspace{1mm}
\noindent\textbf{Training Details.}
During policy learning, we observe that optimizing for both accuracy and efficiency is not effective with a randomly initialized policy. Thus, we fix the policy network and \enquote{warm up} the recognition network using the unimodality models (trained with ImageNet weights) for $5$ epochs to provide a good starting point for policy learning.
We then alternatively train both policy and recognition networks for $20$ epochs and then fine-tune the recognition network with a fixed policy network for another $10$ epochs.
We use same initialization and total number of training epochs for all the baselines (including our approach) for a fair comparison.
We use $5$ segments from a video during training in all our experiments ($C$ set to $5$).
We use Adam~\cite{kingma2014adam} for the policy network and SGD~\cite{sutskever2013importance} for the recognition network following~\cite{wu2019fbnet,sun2019adashare}. We set the initial temperature $\tau$ to $5$, and gradually anneal down to $0$ during the training, as in~\cite{jang2016categorical}.
Furthermore, at test time, we use the same temperature $\tau$ that corresponded to the training epoch in the annealing schedule.
The weight decay is set to $0.0001$ and momentum in SGD is $0.9$. 
$\lambda_k$ is set to the ratio of the computational load between modalities and $\gamma$ is $10$.
More implementation details and source codes are included in the appendix. We will make our source codes and models publicly available.

\vspace{1mm}
\noindent\textbf{Baselines.} We compare our approach with the following baselines and existing approaches. First, we consider unimodality baselines where we train recognition models using each modality separately. Second, we compare with a joint training baseline, denoted as \enquote{Weighted Fusion}, that simply uses all the modalities (instead of selecting optimal modalities per input) via late fusion with learnable weights. This serves as a very strong baseline for classification, at the cost of heavy computation. Finally, we compare our method with existing efficient video recognition approaches, including FrameGlimpse~\cite{yeung2016end}, FastForward~\cite{fan2018watching}, AdaFrame~\cite{wu2019adaframe}, LiteEval~\cite{wu2019liteeval} and ListenToLook~\cite{gao2020listentolook}. 
We directly quote the numbers reported in published papers when possible and use author's provided source codes for LiteEval on both Kinetics-Sounds and Mini-Sports1M datasets.

\vspace{1mm}
\noindent\textbf{Evaluation Metrics.} We compute either video-level mAP (mean average precision) or top-1 accuracy (average predictions of $10$ $224\times224$ center-cropped and uniformly sampled segments) to measure the overall performance of different methods. We also report the average selection rate, computed as the percentage of total segments within a modality that are selected by the policy network in the test set, to show adaptive modality selection in our proposed approach.   
We measure computational cost with giga floating-point operations (GFLOPs), which is a hardware independent metric.

\begin{table}[tb]
\centering
\begin{adjustbox}{max width=\linewidth}
\begin{tabular}{ c|c|c || c|c}
\toprule
 & \multicolumn{2}{c || }{ActivityNet} & \multicolumn{2}{c}{FCVID}  \\
 \cmidrule{2-5}
 
Method & mAP (\%) & GFLOPs  & mAP (\%) & GFLOPs \\
\midrule
FrameGlimpse 
&  60.14 & 33.33 & 67.55 & 30.10\\
FastForward
&  54.64 & 17.86 & 71.21 & 66.11 \\ 
AdaFrame
&  71.5  & 78.69 & 80.2 & 75.13 \\
LiteEval
&  72.7  & 95.1  & 80.0 & 94.3  \\ 
\ours                         & \textbf{73.91} & 94.01 & \textbf{85.82} & 93.86 \\  
\bottomrule
\end{tabular}
\end{adjustbox} \vspace{-1mm}
\caption{\small \textbf{Comparison with state-of-the-art methods on ActivityNet and FCVID}. \ours outperforms LiteEval~\cite{wu2019liteeval} in terms of accuracy ($\sim$$1\%$--$5\%$) with similar computation on both datasets.}
\label{tab:sota}
\vspace{-1pt}
\end{table}
\begin{table}[tb]
\centering
\begin{adjustbox}{max width=\linewidth}
\begin{tabular}{ c|c|c ||  c|c}
 \toprule
 & \multicolumn{2}{c || }{Kinetics-Sounds} & \multicolumn{2}{c}{Mini-Sports1M}  \\
 \cmidrule{2-5}
Method & Acc. (\%) & GFLOPs  & mAP (\%) & GFLOPs \\
\midrule
LiteEval
& 72.02 & 104.06 & 43.64 & 151.83 \\ 
\ours        & \textbf{88.17} & 76.45 & \textbf{46.08} & 138.32 \\  
\bottomrule
\end{tabular}
\end{adjustbox} \vspace{-1mm}
\caption{\small \textbf{Comparison with LiteEval~\cite{wu2019liteeval} on Kinetics-Sounds and Mini-Sports1M}. \ours outperforms LiteEval by a significant margin in both accuracy and GFLOPs on both datasets.} 
\label{tab:liteeval}
\vspace{-1pt}
\end{table}

\subsection{Main Results}

\noindent\textbf{Comparison with Weighted Fusion Baseline.} We first compare \ours with the unimodality and weighted fusion baseline on Kinetics-Sounds and ActivityNet dataset under different task combinations (Table~\ref{tab:afm}-\ref{tab:rgb_audio_flow_k}).
Note that our approach is not entirely focused on accuracy. In fact, our main objective is to achieve both competitive performance and efficiency required for video recognition. As for efficient recognition, it is very challenging to achieve improvements in both accuracy and efficiency. However, as shown in Table~\ref{tab:afm}, \ours outperforms the weighted fusion baseline while offering $47.3\%$ and $35.2\%$ reduction in GFLOPs, on Kinetics-Sounds and ActivityNet, respectively. Interestingly on ActivityNet, while performance of the weighted fusion baseline is worse than the best single stream model (i.e., RGB only), our approach outperforms the best single stream model on both datasets by adaptively selecting input modalities that are relevant for the recognition of the action class.  

Table~\ref{tab:rgb_flow_k} and Table~\ref{tab:rgb_audio_flow_k} show the results of RGB + Flow and RGB + Flow + Audio combinations on the Kinetics-Sounds. Overall, \ours-\texttt{Flow} (which uses optical flow in policy network) outperforms the joint training baseline while offering $50.3\%$ ($304.75$ vs $151.54$) and $56.9\%$ ($308.56$ vs $132.94$) reduction in GFLOPs on RGB + Flow and RGB + Flow + Audio combinations, respectively. \ours-\texttt{RGBDiff} (that uses RGBDiff in policy learning) achieves similar performance compared to \ours-\texttt{Flow} while alleviating computational overhead of computing optical flow (for irrelevant video segments), which shows that RGBDiff is in fact a good proxy for predicting on-demand flow computation during test time. 
In summary, our consistent improvements in accuracy over the weighted fusion baseline with $35\%-55\%$ computational savings, shows the importance of adaptive modality selection for efficient video recognition.

\vspace{1mm}
\noindent\textbf{Comparison with State-of-the-art Methods.}
Table~\ref{tab:sota} shows that \ours outperforms all the compared methods to achieve the best performance of $73.91\%$ and $85.82\%$ in mAP on ActivityNet and FCVID respectively. Our approach achieves $1.21\%$ and $5.82\%$ improvement in mAP over LiteEval~\cite{wu2019liteeval} with similar GFLOPs on ActivityNet and FCVID respectively. Moreover, \ours 
(tested using $5$ segments) outperforms LiteEval by $2.70\%$ ($80.0$ vs $82.70$) in mAP, while saving $39.2\%$ in GFLOPs ($94.3$ vs $57.3$) on FCVID dataset. 
\begin{table}[!tb]
\centering
\begin{adjustbox}{max width=\linewidth}
\begin{tabular}{ c|c|c|c|c}
\toprule
 & \multicolumn{2}{c|}{Network} &  & \\
Method & RGB & Audio   &  mAP (\%) & GFLOPs \\
\midrule
ListenToLook
& ResNet-18 & ResNet-18  & 76.61 & 112.65  \\ 
\ours        & ResNet-18 & ResNet-18 &  80.05 & 82.33 \\  
\midrule
\ours        & ResNet-50 & MobileNetV2 & 84.73 & 110.14 \\  
\ours        & EfficientNet-b3 & EfficientNet-b0 & 85.62 & 30.55 \\  
\bottomrule
\end{tabular} \vspace{-1mm}
\end{adjustbox} \caption{\small \textbf{Comparison with ListenToLook~\cite{gao2020listentolook} on ActivityNet}. \ours outperforms ListenToLook by $3.44\%$ in mAP while offering 26.9\% computational savings in terms of GFLOPs. 
} 
\label{table:listentolook}
\vspace{-1pt}
\end{table}
Table~\ref{tab:liteeval} further shows that \ours significantly outperforms LiteEval by $16.15\%$ and $2.44\%$, while reducing GFLOPS by $26.5\%$ and $8.6\%$, on Kinetics-Sounds and Mini-Sports1M respectively. 
In summary, \ours is clearly better than LiteEval in terms of both accuracy and computational cost on all datasets, making it suitable for efficient recognition.  
Note that FrameGlimpse~\cite{yeung2016end}, FastForward~\cite{fan2018watching} and AdaFrame~\cite{wu2019adaframe} have less computation as they require access to future frames unlike LiteEval and \ours that makes decision based on the current time stamp only.

In addition, we also compare with ListenToLook~\cite{gao2020listentolook} that uses both RGB and Audio streams to eliminate video redundancies. As ListenToLook utilizes weight distillation from Kinetics400 pretrained model, we use Kinetics400 pretrained weights instead of ImageNet weights to initialize our unimodality models for a comparison on ActivityNet in Table~\ref{table:listentolook}. With the same network architecture (ResNet18), \ours outperforms ListenToLook by a margin of $3.44\%$ in mAP while using $26.9\%$ less computation. 
This once again shows that our proposed approach of adaptively selecting right modalities on a per segment basis is able to yield not only significant savings in computation but also improvement in accuracy.
To show that the benefits of our method extend even to more recent and efficient networks, we use EfficientNet~\cite{tan2019efficientnet} in our approach and observe that it provides the best recognition performance of $85.62\%$ in mAP with only $30.55$ GFLOPs ($\sim$$73\%$ less computation compared to \ours(ResNet50 | MobileNetV2)).

\begin{table}[t]
\centering 
\begin{adjustbox}{max width=\linewidth}
\begin{tabular}{ c | cc | cc | cc}
 \toprule
   \cmidrule{1-7}
 & \multicolumn{2}{c|}{RGB + Audio} & \multicolumn{2}{c|}{RGB + Flow} & \multicolumn{2}{c}{RGB + Flow + Audio}  \\
 \cmidrule{2-7}

Method  & Acc. (\%) & GFLOPs & Acc. (\%) & GFLOPs & Acc. (\%) & GFLOPs\\
\midrule
Average Fusion  & 88.15 & 145.17 & 83.30  & 304.75 & 88.18  & 308.56 \\
Class-wise Weighted Fusion  & 87.86 & 145.17 & 83.82 & 304.75 & 87.75 & 308.56 \\
Max Fusion  & 86.49 & 145.17 & 83.47 & 304.75 & 88.06 & 308.56\\
FC2 Fusion$^*$  & 87.73 & 145.17 & 83.30 & 304.75 & 87.84 & 308.56 \\
Weighted Fusion & 87.86 & 145.17 & 83.47 & 304.75 & 88.25 & 308.56\\
\ours & \textbf{88.17} & \textbf{76.45} & \textbf{84.36}  & \textbf{137.03} & \textbf{89.06}  & \textbf{141.97} \\
\bottomrule
\multicolumn{7}{l}{\footnotesize{$^*$: concatenating feature vectors from all modalities and add two addition fully-connected layers to fuse features.}}
\end{tabular}
\end{adjustbox} \vspace{-1mm}
\caption{\small \textbf{Comparison with fusion strategies on Kinetics-Sounds}. \ours consistently outperforms hand-designed fusion strategies with overall $50\%-60\%$ computational savings.} 
\label{tab:diff_fusion} \vspace{-1mm}
\end{table}

\subsection{Ablation Studies}

\noindent\textbf{Comparison with Additional Fusion Strategies.} We compare with four additional fusion strategies including weighted fusion on different combinations of modalities. Table~\ref{tab:diff_fusion} shows that our approach \ours consistently outperforms all the hand-designed fusion strategies while offering $47.3\%$, $55.03\%$ and $53.99\%$ reduction in GFLOPs on RGB + Audio, RGB + Flow and RGB + Flow + Audio combinations on Kinetics-Sounds respectively. Furthermore, \ours with RGBDiff as the proxy alleviates the computational overhead of computing optical flow (which is often very expensive) making it suitable in online scenarios. Similarly, \ours offers $19.2\%$ computational savings while outperforming these fusion strategies by a margin of about $2\%$ in mAP on RGB + Audio combination on ActivityNet. 

\vspace{1mm}
\noindent\textbf{Policy Design.} We investigate the effectiveness of our design of policy by either selecting or skipping both modalities at same time instead of taking the decisions per modality. In other words, we use a single FC layer in the policy network which outputs the binary decisions where 1 indicate the use of both modalities and 0 indicate skipping of both modalities in our framework. \ours outperforms the alternative design ($88.02$ vs $88.17$) while saving $18\%$ GFLOPS on Kinetics-Sounds. Selection of both modalities at the same time increases the computation as it favors selection of more RGB stream. On the other hand, \ours selects comparatively less RGB stream and focuses more on the cheaper audio stream as many actions can be recognized by only audio without looking into the RGB frames.

\begin{table} 
\centering
\begin{adjustbox}{max width=\linewidth}
\begin{tabular}{ c|  c ||  c}
 \toprule
 & AcitivityNet & Kinetics-Sounds   \\
 \cmidrule{2-3}
Method & mAP (\%) &  Acc. (\%)\\
\midrule
Random (Train) & 70.34 & 84.34\\
Random (Test)  & 58.31 & 85.75 \\
Random (Train + Test) & 70.85 & 86.28 \\
\ours  & \textbf{73.91} & \textbf{88.17} \\
\bottomrule
\end{tabular} 
\end{adjustbox} \vspace{-1mm}
\caption{\small \textbf{Comparison with random policy on RGB + Audio}. Random (X) denotes random selection of modalities during X-phase of the learning. \ours outperforms all the variants showing effectiveness of learned policy in video recognition.} \label{tab:random}
\vspace{-1pt}
\end{table}
\vspace{1mm}
\noindent\textbf{Comparison with Random Policy.} We perform three different experiments by randomly selecting a modality with $50\%$ probability during both training and testing. Table~\ref{tab:random} shows that our approach \ours outperforms all the three variants by a large margin (e.g., $15.60\%$ and $2.42\%$ improvement over Random (Test) on ActivityNet and Kinetics-Sounds respectively) which demonstrates the effectiveness of our learned policy in selecting the optimal modalities per input instance while recognizing videos. 

\begin{figure*}[bt]
    \centering
     \includegraphics[width=1\linewidth]{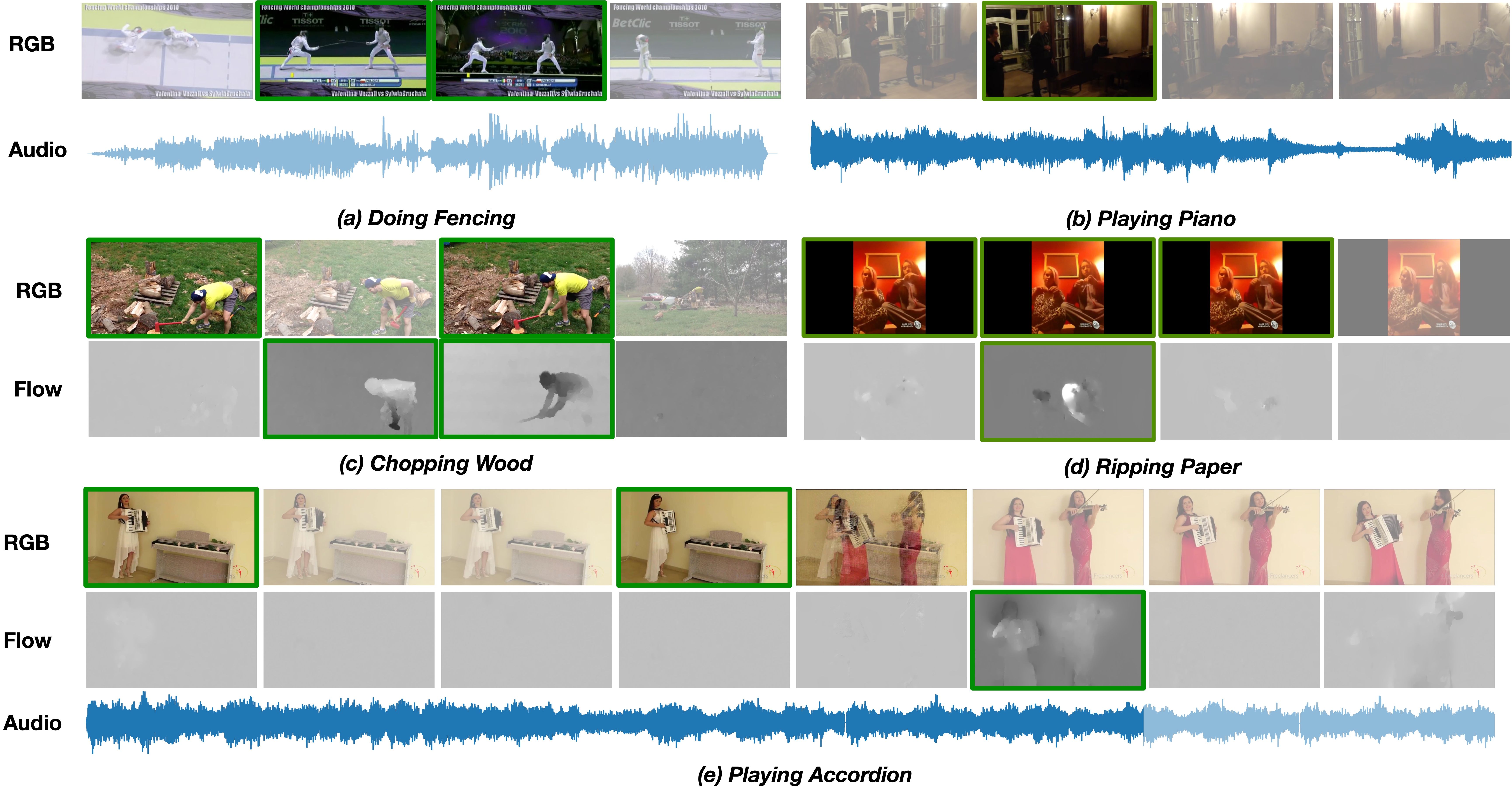}
    \vspace{-6mm}
   \caption{
  \small 
  \textbf{Qualitative examples showing the effectiveness of \ours in selecting the right modalities per video segment (marked by green borders).} 
  (a, b) RGB + Audio: \ours selects RGB stream for second and third segments in (a) while skips irrelevant audio coming from the reporter and background song. Similarly in (b), it is able to select RGB modality for only one segment while selecting the entire audio stream as the action can be easily recognized with audio (Playing Piano).
  (c, d) RGB + Flow: Our approach selects flow stream only when it is informative for the action, e.g., second and third segments in (c) and only second segment in (d).
  (e) RGB + Flow + Audio: \ours selects audio for most of the segments (not for last two segments as the audio is not clear with the mixing of sound from both instruments) while selecting flow only for the sixth segment where the motion related to the action is clearly visible. Best viewed in color.}
  \vspace{-3pt}
   \label{fig:qual}
\end{figure*}

\vspace{1mm}
\noindent\textbf{Ablation on Training Losses.} 
As discussed in Section~\ref{sec:loss}, $\lambda_k$ and $\gamma$ controls the trade-off between accuracy and computational efficiency. 
We investigate the effect of efficiency loss in RGB + Audio experiment on Kinetics-Sounds and observe that training without efficiency loss (both $\lambda_k$ and $\gamma$ set to $0$) achieves a video accuracy of $88.82\%$ (an improvement of $0.65\%$) while requiring $47.3\%$ more computation than \ours that uses efficiency loss during training. Similarly, using equal cost weights for both modalities (by setting $\lambda_{rgb}$=$\lambda_{audio}$=1) achieves an accuracy of $86.82\%$ compared to $88.17\%$ using \ours, with very less utilization of audio (only $39.13\%$ in contrast to $94.15\%$ using our approach). As processing audio stream is much cheaper, we use $\lambda_{rgb}=1$ and $\lambda_{audio}=0.05$ to favor selection of cheaper modalities and achieves an accuracy of $88.17\%$ with $76.45$ GFLOPs on Kinetics-Sounds. We further test the effect of penalty factor $\gamma$ in Equation~\ref{eq:loss} by varying it from [$0$, $2$, $5$, $10$] and observe that it has little effect on the final performance with the best performance at $\gamma=10$ in all our experiments. 

\vspace{1mm}
\noindent\textbf{Effectiveness of LSTM.} We investigate the effectiveness of LSTM in modeling video causality on the RGB + Audio experiment and observe that directly predicting a choice via a single fully-connected layer (i.e., by removing LSTM from the policy network) decreases the video accuracy from $88.17\%$ to $86.82\%$ on Kinetics-Sounds. This confirms that LSTM is critical for good performance as it makes the policy network aware of all useful information seen so far.  

\vspace{1mm}
\noindent\textbf{Sampling Hyperparameters.} We test the effect of temperature (Equation~\ref{eq:one}) in RGB+Audio experiment on Kinetics-Sounds dataset by varying it from [$0$, $0.5$, $5$, $10$] and observe that higher values ($5$, $10$) show better performance (by $0.5\%$-$0.7\%$) than lower ones. So, we start at a high temperature (set to $5$ in all our experiments) and anneal it to a small non-zero value, as in~\cite{jang2016categorical}. Similarly, we also vary the annealing factor from [$0$, $0.5$, $0.965$] and notice that setting it to $0.965$ leads to the best accuracy of $88.17\%$ while $0$ leads to an accuracy of $87.40\%$ on Kinetics-Sounds.

\subsection{Qualitative Results}
Figure~\ref{fig:qual} shows the selected modalities using our approach on different cases ((a, b) RGB + Audio, (c, d) RGB + Flow, and (e) RGB + Flow + Audio). 
As seen from Figure~\ref{fig:qual}.(a), our approach is able to select RGB modality for the segments that are more informative of the action and skip the audio stream as audio in that video is irrelevant to the action \enquote{fencing} (majority of audio comes from the reporter and background song). Similarly in Figure~\ref{fig:qual}.(b), it is able to select RGB modality for only one segment while selecting the entire audio stream as the action can be easily recognized with audio (\enquote{Playing Piano}). Overall, we observe that \ours focuses on the right modalities to use per segment for correctly classifying videos while taking efficiency into account (e.g., in Figure~\ref{fig:qual}.(e), it mainly focus on audio for most of the segments while selecting RGB only for two informative segments and flow stream for the sixth segment for recognizing the action \enquote{Playing Accordion}).

\section{Conclusion}
\label{sec:conclusion}

In this paper, we present \ours, a novel and differentiable approach for adaptively determining what modalities to use per segment per instance for efficient video recognition. 
In particular, we trained a multi-modal policy network to predict these decisions with the goal of achieving both competitive accuracy and efficiency. 
We efficiently train the policy network jointly with the recognition model using standard back-propagation.
We demonstrate the effectiveness of our proposed approach on four standard datasets, outperforming several competing methods.

{\small
\bibliographystyle{ieee_fullname}
\bibliography{egbib}
}

\clearpage
\appendix

\vspace{1mm}
\section{Dataset Details} We evaluate the performance of our approach using four standard video datasets, namely ActivityNet-v1.3~\cite{caba2015activitynet}, FCVID~\cite{jiang2017exploiting}, Mini-Sports1M~\cite{karpathy2014large} and Kinetics-Sounds~\cite{arandjelovic2017look}. Below we provide more details on each of the dataset.

\vspace{1mm}
\noindent\textbf{ActivityNet.} We use the v1.3 split which consists of more than 648 hours of untrimmed videos from a total of 20K videos. Specifically, this dataset has 10,024 videos for training, 4926 videos for validation
and 5044 videos for testing with an average duration of 117 seconds. It contains 200 different daily activities such as: walking the dog, long
jump, and vacuuming floor. As in literature, we use the training videos to train our network, and the validation set for testing as labels in the testing set are withheld by the authors. The dataset is publicly available to download at \url{http://activity-net.org/download.html}.

\vspace{1mm}
\noindent\textbf{FCVID.} Fudan-Columbia Video Dataset (FCVID) contains total 91,223 Web videos annotated manually according to 239 categories (45,611 videos for training and 45,612 videos for testing).
The categories cover a wide range of topics like social events, procedural events, objects, scenes, etc. that form in a hierarchy of 11 high-level groups (183 classes are related to events and 56 are objects, scenes, etc.). The total duration of FCVID is 4,232 hours with an average video duration of 167 seconds. The dataset is available to download at \url{http://bigvid.fudan.edu.cn/FCVID/}. 

\vspace{1mm}
\noindent\textbf{Mini-Sports1M.} Mini-Sports1M is a subset of Sports-1M~\cite{karpathy2014large} dataset with 1.1M videos of 487 different fine-grained sports. It is assembled by~\cite{gao2020listentolook} using videos of length 2-5 mins, and randomly sample 30 videos for each class for training, and 10 videos for each class for testing. The classes are arranged in a manually-curated taxonomy that contains internal nodes such as Aquatic Sports, Team Sports, Winter Sports, Ball Sports, etc, and generally becomes fine-grained by the leaf level. We obtain the training and testing splits from the authors of~\cite{gao2020listentolook} to perform our experiments. Both training and testing videos in this dataset are untrimmed. This dataset is available to download at \url{https://github.com/gtoderici/sports-1m-dataset}.

\vspace{1mm}
\noindent\textbf{Kinetics-Sounds.} Kinetics-Sounds (assembled by~\cite{arandjelovic2017look}) is a subset of Kinetics and consists of 22,521 videos for training and 1,532 videos testing across 31 action classes. The original subset contains 34 classes, which have been chosen to be potentially manifested visually and aurally, such as playing various instruments (guitar, violin, xylophone, etc.), using tools (lawn mowing, shovelling snow, etc.), as well as performing miscellaneous actions (tap dancing, bowling, laughing, singing, blowing nose, etc.). Since 3 classes were removed from the original Kinetics dataset, we use the remaining 31 classes in our experiments, as in~\cite{gao2020listentolook}. Although this dataset is fairly clean by construction, it still contains considerable noise and many videos contain sound tracks that are completely unrelated to the
visual content (e.g. Doing fencing in Figure 3.(a) of the main paper) which makes it suitable for our approach to adaptively select right modalities conditioned on the input. The original Kinetics dataset is publicly available to download at \url{https://deepmind.com/research/open-source/kinetics} and the classes for Kinetics-Sounds can be obtained from~\cite{gao2020listentolook}.

\section{Implementation Details}

For our experiments, we use 12 NVIDIA Tesla V100 GPUs for the RGB + Audio experiments and 18 GPUs for both RGB + Flow and RGB + Flow + Audio experiments. All our models were implemented and trained via PyTorch.

\vspace{1mm}
\noindent\textbf{Network.}
For non-audio modality, we add temporal max-pooling layers (kernel size 3, stride 2) to reduce computations. In recognition network, we use TSN-like ResNet-50 network~\cite{wang2016temporal} with three temporal max-pooling layers which are located at the beginning of stage 2, 3 and 4 of ResNet-50 (there are 4 stages in ResNet-50), i.e., the third, forth and fifth locations of reducing spatial resolution in the network.
On the other hand, we add two temporal max-pooling layers to the MobileNetV2 used in the policy network for non-audio modality since the number input frames for policy network is fewer compared to the recognition network.

\vspace{1mm}
\noindent\textbf{Input.}
We first use FFMPEG to extract RGB frames and Audio from a video. While decoding a video into RGB frames, the shorter side of the RGB frames is resized to 256 while keeping the aspect ratio. We use the resized frames to compute optical flow via TV-L1 algorithm and bound the flow range to [-20, 20]. We convert the audio to single-channel and resample it at 24kHz. 
During training, we divide a video into $C$ equal-length regions ($C=5$ in our experiments). For each region, we randomly pick 32 consecutive frames and uniformly subsample 8 frames as a RGB segment, i.e., the temporal stride between frames is 4.
For the Audio data, we take a 1.28s-length window that is center-aligned to the RGB frames and then we use short-time Fourier transform to convert the audio into a log-spectrogram of window length 10ms, hop length 5ms with 256 frequency bins~\cite{kazakos2019epic}. 
For the Flow data, at each RGB frame location, we stack horizontal and vertical flow of 5 consecutive frames interleavedly to form the input. Moreover, for the frame difference used in the multi-modal policy network, we follow the same practice as in optical flow, and stack 5 consecutive frame difference images to form the input~\cite{wang2016temporal}. On the other hand, $C$ is 10 during testing as we use 10 video segments.

\vspace{1mm}
\noindent\textbf{Training.}
We use a batch size of 72 with synchronized batch normalization in all our experiments, 
The data augmentations for the RGB and Flow modalities are based on the practices in~\cite{wang2018non}. We first randomly resize the shorter side of an image to a range of [256, 320) while keeping aspect ratio and then randomly crop a $224\times224$ region and normalize it with the ImageNet's mean and standard deviation to form the input ($8\times224\times224$). For the Audio modality, we simply take the $256\times256$ spectrogram as the input. The same data augmentations are used in the policy network while the data of the non-audio modality is further downsampled in both temporal and spatial dimension ($4\times160\times160$). 
The training time depends on the size of datasets and the task. E.g., for the RGB + Audio task, it takes about 12 hours for Kinetics-Sound and 16 hours for ActivityNet.

\vspace{1mm}
\noindent\textbf{Testing.}
During testing, we uniformly sample 10 video segments from a video. For RGB and Flow modalities, we resize the shorter side of an image to 256, and then crop a center $224\times224$ region for evaluation.

\section{Discussion on RGB Difference} As described in Section 4 of the main paper, we utilize RGB frame difference as a proxy to optical flow in our policy network and compute flow when needed since computing flow is very expensive. Here we compare RGBDiff and flow in terms of unimodal and weighted fusion (joint learning) when combined with RGB performance to further verify the effectiveness of RGBDiff on Kinetics-Sounds. Table~\ref{tab:rgbdiff} shows that RGBDiff outperforms Flow in unimodal performance (75.73\% vs 80.10\%) whereas both Flow and and RGBDiff (when combined with RGB) performs very similar (83.47 vs 83.30) on Kinetics-Sounds. This shows that RGBDiff as an input modality is also quite effective both in unimodal and joint learning performance and hence can be used as a proxy in the policy network for predicting the on-demand flow computation during test time.

\begin{table}[t]
\centering
\begin{adjustbox}{max width=0.8\linewidth}
\begin{tabular}{ c|c|c}
\toprule
Method &  Acc. (\%)  & GFLOPs   \\
\midrule
RGB       & 82.85  &  141.36  \\
Flow      & 75.73 &  163.39  \\
RGBDiff   & 80.10 & 179.12   \\
Weighted Fusion (RGB + Flow)  & 83.47 & 304.75  \\
Weighted Fusion (RGB + RGBDiff) & 83.30 & 320.48 \\
\bottomrule
\end{tabular}
\end{adjustbox} \vspace{-1mm}
\caption{\small {Comparison between Optical Flow and RGB Difference on Kinetics-Sounds}. RGBDiff as an input modality is very competitive with optical flow in both unimodal and joint learning.} \vspace{-3mm}
\label{tab:rgbdiff}
\end{table}

\section{Qualitative Results}

\begin{figure}[t]
    \centering
     \includegraphics[width=1\linewidth]{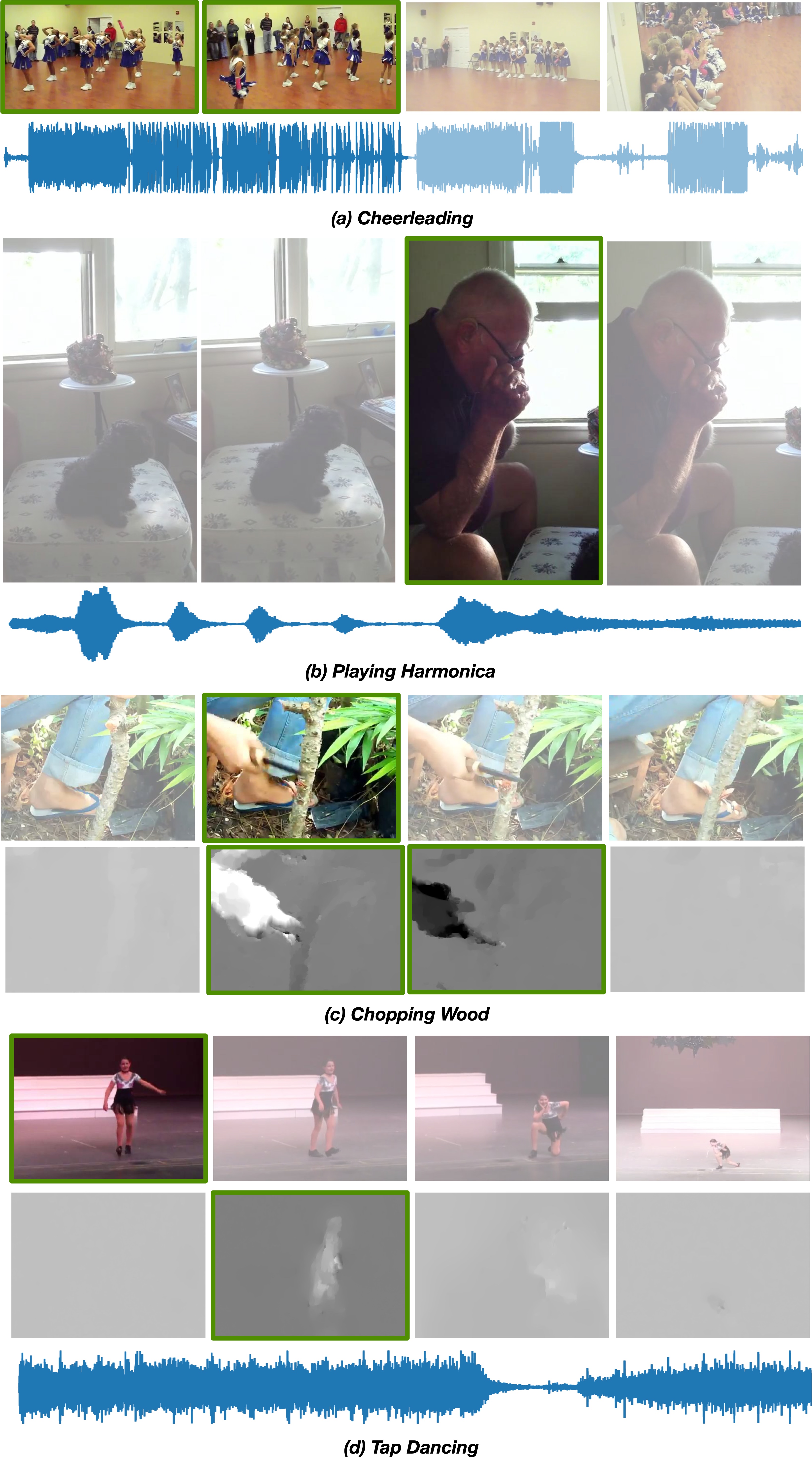} \vspace{-1mm}
   \caption{\small 
  \textbf{More qualitative examples showing effectiveness of \ours in selecting right modalities per video segment (marked by green borders).} Overall, we observe that our approach focuses on the right modalities to use per segment for correctly classifying the videos while taking efficiency into account. 
  }
   \label{fig:more_qual} \vspace{-1mm}
\end{figure}

Figure~\ref{fig:more_qual} shows the selected modalities using our approach on different cases. 
As seen from Figure~\ref{fig:more_qual}.(a), 
our approach selects relevant RGB and audio for only first two segments as both modalities become irrelevant for last two segments as girls are discussing instead of cheerleading. Similarly, in Figure~\ref{fig:more_qual}.(b), \ours is able to select RGB for only one segment that is more informative of the action and selects the entire audio stream as the action can be easily recognized with audio (Playing Harmonica). Figure~\ref{fig:more_qual}.(c) and (d) shows two more examples of RGB + Flow and RGB + Flow + Audio experiments respectively, where our approach selects the right modalities to use per segment (e.g., in Figure~\ref{fig:more_qual}.(d), it mainly focuses on audio while selecting RGB and flow for only two segments) for correctly classifying the videos while taking efficiency into account. 

\end{document}